\documentclass{article}

\usepackage{arxiv}

\usepackage[utf8]{inputenc} 
\usepackage[T1]{fontenc}    
\usepackage{hyperref}       
\usepackage{url}            
\usepackage{booktabs}       
\usepackage{amsfonts}       
\usepackage{nicefrac}       
\usepackage{microtype}      
\usepackage{lipsum}		
\usepackage{graphicx}
\usepackage[sort,numbers]{natbib}
\usepackage{doi}

\usepackage{svg}
\graphicspath{{./img/}} 
\usepackage{amssymb} 
\usepackage{amsmath}
\usepackage{subfigure}

\title{Noise-Resilient Ensemble Learning using Evidence Accumulation}
\author{
	\href{https://orcid.org/0000-0002-9072-1535}{\includegraphics[scale=0.06]{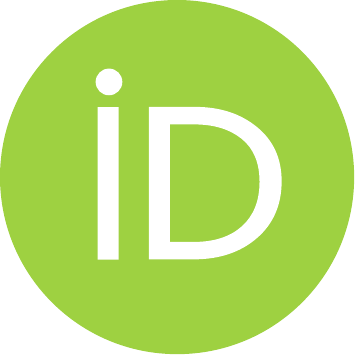}\hspace{1mm}Ga\"elle Candel} \\
	Wordline TSS Labs, Paris \\
	\texttt{firstname.lastname@worldline.com} \\
	\& \\
	D\'epartement d'informatique de l'ENS \\
	ENS, CNRS, PSL University, Paris \\
	\texttt{firstname.lastname@ens.fr}\\
	\And
  David Naccache \\
	D\'epartement d'informatique de l'ENS \\
	ENS, CNRS, PSL University, Paris \\
	\texttt{firstname.lastname@ens.fr}\\
}

\date{} 					

\renewcommand{\headeright}{}
\renewcommand{\undertitle}{}

\renewcommand{\shorttitle}{Noise-Resilient EL using EAC}

\hypersetup{
pdftitle={Noise-Resilient Ensemble Learning using Evidence Accumulation},
pdfsubject={cs.AI},
pdfauthor={Ga\"elle Candel, David Naccache},
pdfkeywords={Classification, Clustering, Distributed systems, Decentralized systems, Ensemble Learning, Evidence accumulation clustering, Label corruption, Peer-to-Peer},
}

\begin{document}
\maketitle

\begin{abstract}
	\textit{Ensemble Learning} methods combine multiple algorithms performing the same task to build a group with superior quality.
	These systems are well adapted to the distributed setup, where each peer or machine of the network hosts one algorithm and communicate its results to its peers.
	Ensemble learning methods are naturally resilient to the absence of several peers thanks to the ensemble redundancy.
	However, the network can be corrupted, altering the prediction accuracy of a peer, which has a deleterious effect on the ensemble quality.
	In this paper, we propose a noise-resilient ensemble classification method, which helps to improve accuracy and correct random errors.
	The approach is inspired by \textit{Evidence Accumulation Clustering} , adapted to classification ensembles.
	We compared it to the naive voter model over four multi-class datasets.
	Our model showed a greater resilience, allowing us to recover prediction under a very high noise level.
	In addition as the method is based on the evidence accumulation clustering, our method is highly flexible as it can combines classifiers with different label definitions.

\end{abstract}

\keywords{Classification \and Clustering \and Distributed systems \and Ensemble learning \and Evidence accumulation clustering \and Label corruption \and Noise resilience \and Peer-to-Peer}

\section{Introduction}

\textit{Ensemble Learning} \cite{Zhou2012EnsembleMF} methods combine several algorithms performing the same task to obtain a better-quality group.
Ensemble learning methods play on diverse group aspects: the number of algorithms \cite{Breiman2004BaggingP,Khan2020EnsembleOO}, their weighting based on their contribution \cite{Schapire2012BoostingFA,Li2008WeightedCC,Ren2013WeightedObjectEC}, and their selection based on their diversity \cite{Margineantu1997PruningAB}.

Ensemble learning methods are well adapted to the distributed setup, where several machines host each a single algorithm and send their results to a central node aggregating the results \cite{Fan1999TheAO,Ormndi2013GossipLW,Abualkibash2013HighlySP}.
They can be adapted to decentralized peer-to-peer networks \cite{Ormndi2013GossipLW}, where a dynamic group collaborate to improve its accuracy by electing a leader or by aggregating the group's results.
Distributed systems are prone to network failures, where communications are broken between some nodes, or corrupted with noise \cite{Ratasich2019ART}.
In addition, nodes can change of behavior if controled by malicious entities.

Ensemble methods are resilient to the absence of one or more weak learners thanks to group redundancy \cite{Margineantu1997PruningAB,Probst2017ToTO}.
However, the corruption of a learner's predictions is equivalent to a negative change of accuracy, which has a deleterious effect on the group quality.
Thus, there are two ways to deal with corrupted computers: detecting inaccurate peers to avoid data pollution or resilience to error.
The detection can be done using network monitoring methods, or exploiting trust to weigh peers based on their past contributions.
However, this approach is not adapted to a dynamic environment such as a peer-to-peer network where a peer lifetime is very short and may change temporary behavior.
In contrast, being resilient to error is more suitable as all inputs are accepted but more challenging to design as it requires smart correction algorithms.

In this article, we propose a noise-resilient ensemble classification method, correcting errors while improving accuracy.
The method uses the Evidence Accumulation Clustering approach to rectify class boundaries and correct corrupted labels by performing a local weighted vote.
The approach was tested under several noise condition over four datasets and tolerated high noise levels without accuracy degradation.

This paper is structured as follows.
The first section presents the related works regarding ensemble learning methods and resilience to error.
The second section details the proposed ensemble classification method.
The datasets and the classifiers' setup are detailed in the experimental section, followed by the results.
Finally, the paper ends with a discussion and a conclusion.

\section{Related Works}

Ensemble classification  methods adopt different strategies to increase accuracy.
The simplest one is \textit{bagging} \cite{Breiman2004BaggingP} -- for bootstrap aggregating, learners are trained over randomly sampled subsets.
A related approach exploits \textit{random projections} \cite{Khan2020EnsembleOO}, which creates a different view of the data, making it less dependent on the pre-processing step.
Rather than sampling items at random, selection can be made on features \cite{Pes2019EnsembleFS}, reducing the computational complexity of each classifier while allowing the ensemble to classify incomplete items with missing features.
A classifier often makes very few errors in very dense areas where a single class is represented,  because there is no ambiguity.
Near a boundary, the classifier is less exact because this area is less dense; therefore difficult to learn correctly.
Ensemble methods help mostly to correct these areas.
A good example is decision trees with rough decision boundaries, while a \textit{Random Forest} \cite{breiman2001random} has smooth boundaries with various shapes.
The \textit{bagging} approaches exploit the fact that random initialization would lead to diverse classifiers, compensating errors within the group.
However, it requires a large number of classifiers and fails to improve when classifiers are correlated.

\textit{Boosting} \cite{Schapire2012BoostingFA} solves the correlation problem by weighting classifiers based on the improvement they can lead to.
On the other hand, \textit{ensemble pruning} \cite{Margineantu1997PruningAB} removes correlated classifiers that lead to no improvement, decreasing the overall complexity while preserving the accuracy.
The ensemble size is smaller for these methods but is still resilient to the absence of several classifiers during the inference stage \cite{Probst2017ToTO}.
\textit{Boosting} and \textit{ensemble pruning} target the problem of \textit{how to construct} a full ensemble using very few classifiers.
As they require a setup phase to evaluate the classifiers, these approaches are not adapted to dynamic environments, such as P2P networks, as peers can join and leave at any time.

Most ensemble classification methods assume that no issue can occur within the ensemble, i.e. all predictions are transmitted without errors.
The centralizing node is always available, but some classification peers can be down.
In the case of Internet-of-Things networks, like Wireless Sensor Network (WSN) or Vehicular Ah Hoc Network (VANET), the devices are prone to fault, failures and attacks \cite{Ratasich2019ART,Sen2009ASO,Sheikh2019ACS}.
Several techniques exist to detect deceptive nodes, but it requires some time and often needs centralized monitoring capabilities.
To our knowledge, no work assumed transmission of corrupted predictions to the aggregator node, nor solution in case of unavailability.
It is equivalent to a completely decentralized P2P network where there is no leader and corrupted peers can participate to the classification task.

The other branch of ensemble learning is \textit{ensemble clustering}  \cite{Strehl2002ClusterE} that combines clustering algorithms.
There is no way to know if a partitioning is \textit{good} because there is no ground truth.
Therefore, ensemble clustering algorithms must simultaneously deal with the good and bad clustering without any other help.
Therefore, this class of algorithms will be helpful to deal with corrupted predictions.
While clustering is analogous to classification as it assigns labels to items,
the approaches shared by ensemble classication and clustering are limited to bagging and weighting \cite{Li2008WeightedCC,Ren2013WeightedObjectEC},
where weights are assigned at \textit{inference time} allowing to handle corrupted output more carefully.


\textit{Evidence Accumulation Clustering}  \cite{Galdi2014ConsensusCI,Fred2005CombiningMC,Li2008WeightedCC,Monti2004ConsensusCA,Ren2013WeightedObjectEC}  is one of the main Ensemble clustering methods, which has the advantage of not requiring to match labels between the different clustering.
The approach starts by gathering the co-clustering frequencies of all possible pairs of items into a co-association (CA) matrix of size $n \times n$ for $n$ items.
The similarity matrix obtained is then re-clustered to obtain the final partitioning.
A hierarchical algorithm is often used \cite{Fred2005CombiningMC,Galdi2014ConsensusCI} as it allows the final user to decide on the clustering granularity.
The major drawback of evidence accumulation clustering is the complexity and the memory footprint because of the $n \times n$ matrix.
Using \textit{Single-Linkage} (SL) hierarchical clustering \cite{Fred2005CombiningMC}, the authors proposed to limit the CA matrix to the $k=20$ nearest neighbors,
as SL does not consider distant neighbors.
Depending on the dataset, the approach led to better results than other clustering methods exploiting the full matrix while reducing the overall computational complexity.

We inspired ourselves from the evidence accumulation clustering methods and the nearest-neighbors trick to deal with noise and improve accuracy.
Nonetheless, our approach differs significantly from it as the CA weights are exploited to refine labels rather than clustered to obtain the final classification.
The approach will be detailed in the following section.

\clearpage

\pagebreak

\section{Label Refinement with Implicit Boundary Learning}

In this section, we will detail our proposed ensemble classification method exploiting the evidence accumulation clustering method.
The task is to classify an unlabeled dataset $X = \{\mathbf{x}_1, \mathbf{x}_2, ..., \mathbf{x}_n\}$ with $n = |X|$ elements.
The ensemble is a group $\mathcal{A} = \left\{A^{(1)}, A^{(2)}, ..., A^{(k_p)}\right\}$ of $k_p$ classifiers, called sometimes \textit{peers}.
The way they are obtained impacts the ensemble accuracy -- as in any other ensemble classification methods -- but does not impact the overall process.
The prediction made by the peer $p$ is denoted $\hat{Y}^{(p)} = A^{(p)}(X) \in \left(\mathcal{L}^{(p)}\right)^n$, where $\mathcal{L}^{(p)}$ is the set of classes that $p$ can distinguish, possibly different from the other classifiers. We will discuss later about this particularity.

The goal of the proposed approach is to rectify the label of an item based on its neighborhood.
In general, a classifier makes errors on items close to a class boundary.
A peer will collect the peers' opinion to know if two items are on the same side of the class boundary.
These weights will be used to rectify uncorrect labels using a weighted voter model.
The following paragraphs describe the process and explain the motivation of the different choices.

\subsection{Gathering Co-Association Matrices}

The initial prediction $\hat{Y}^{(p)}$ of the peer $p$ is transformed into the local CA matrix $M^{(p)}$, where $M^{(p)}(\mathbf{x}, \mathbf{x}') = 1 \text{ if } A^{(p)}(\mathbf{x}) = A^{(p)}(\mathbf{x}') \text{ else } 0$.
Only the pairs $M(\mathbf{x}, \mathbf{x}')$  concerning the $k$-nearest neighbors ($k$-NN) of $\mathbf{x} \in X$ (denoted $\mathcal{N}_k(\mathbf{x})$, with $\mathbf{x} \notin \mathcal{N}_k(\mathbf{x})$) are computed.
As mentioned in \cite{Fred2005CombiningMC}, this trick reduces the memory footprint and computational cost from $\mathcal{O}(n^2)$ to $\mathcal{O}(kn)$.

After prediction, peers exchange their results, allowing them to compute locally the average CA matrix $\mathcal{M}$:
\begin{equation}\label{eq:ca_average}
  \mathcal{M}(\mathbf{x}, \mathbf{x}') = \frac{1}{k_p}\sum_{p=1}^{k_p} M^{(p)}(\mathbf{x}, \mathbf{x}')
\end{equation}
Peers may get different results of $\mathcal{M}$ as a perfect communication model is not assumed here.
Some peers may not receive the results from all of their peers, or may get truncated messages depending on the network stability.
They replace in that case $\frac{1}{k_p}$ by the number of message received for a particular pair $(\mathbf{x}, \mathbf{x}')$.
Unless very few messages are received by a peer, this would not impact the outcome of the process.

\paragraph{Nearest Pairs:}

The computation of $\mathcal{M}$ for the closest pairs only is motivated by the idea that NN are likely to belong to the same class unless near a class boundary.
In a multi-class classification problem, negative information -- indicating that two items belong to different classes -- is less informative than positive information.
For example, for $d$ classes and two items, there are $d \times (d-1)$ assignment possibilities for negative evidence while only $d$ for positive ones.
With that many possibilities, it is unlikely to find the correct classes by chance using negative information.
Additionally, for an item in a class representing $100 \times \alpha \%$ of the dataset, $100 \times (1  - \alpha) \%$ of the pairs would be $0$, leading to many unnecessary data as $\alpha$ shrinks with the number of classes and the class imbalance.

\paragraph{Class Matches:}
The use of pairs of co-association  makes it possible to dispense with the definition of peer classes $\mathcal{L}^{(p)}$.
$\mathcal{M}$ gather the number of times two items are classified together, regardless of the possible labeling errors or granularity level.
The evidence accumulation clustering approach allows combining classification to clustering algorithms as they both produce labeling, one following a ground-truth definition, the other defined from scratch.
Nevertheless, the combination of algorithms with different labels definitions assumes that most of the boundaries are common to the different classes/clusters

As an example, suppose we have three classifiers with different goals,
one classifier focusing on \textit{animals} $[\text{dog}, \text{wolf}, \text{cat}, \text{lion}, \text{frog}]$,
another on \textit{sizes} $[$ small, medium, large$]$,
and the last on \textit{colors} $[\text{white}, \text{gray}, \text{black}, \text{green}]$.
The \textit{animal} and \textit{size} classifiers are compatible as one size corresponds to a unique set of animals  ($\text{small} = \{\text{frog}\}$, $\text{medium} = \{\text{dog}, \text{cat}\}$, $\text{large} = \{\text{wolf}, \text{lion}\}$).
Therefore, all the \textit{size} boundaries are preserved by the \textit{animal} classifiers, with some additional.
However, the \textit{animal} and \textit{color} classifiers share only the boundary between $\text{frog}$ and the other animals which corresponds to the boundary between $\text{green}$ and the other colors.
Therefore, these two classifiers are weakly compatible.

To obtain compatible classifiers, classes must be derived from a primary set of classes,
as in \textit{Error Correcting Output Code} (ECOC) \cite{Dietterich1995SolvingML} where classes are grouped into two groups to train a binary classifier on it.
Another option is available if classes organized into a hierarchical ontology.
In this case,  high and low-level classes could be combined as several low-level classes derived from a single high-level class.
Therefore, the high-level class boundary is preserved within the low-level classes.
However, the combination of algorithms with different class definitions impacts the strength of $\mathcal{M}$ as the ensemble boundaries are less pronounced.
This would have almost no impact on the high-level classifiers as low-level ones share their boundaries.
Still, low-level classifiers would be as not all their boundaries are not  matched by higher-level classifiers.

\paragraph{Exchanging Pairs:}
Depending on the network quality and the data privacy preferences, a peer may prefer to exchange its CA matrix or its raw predictions.

The raw predicitions require $8 \times n$ bits per message if a label is encoded over an octet in terms of \textit{bandwidth requirements}.
When sending the binary CA matrix, it requires $k \times n$ bits per message.
In both cases, we assume that all peers have the same set of items with the same indexing; therefore the features and index do not need to be exchanged.
The value of $k$ in our setups is relatively small ($\approx 10$); therefore one or the other option is equivalent in size.

In an \textit{insecure environment}, peers may prefer not to exchange their raw prediction as an eavesdropper would get roughly labeled data for free.
With a CA matrix, the  eavesdropper can only recover a partial clustering, limiting the amount of information leakage.
In addition, the encryption scheme \cite{Hao2010AnonymousVB} can be used in binary form,
allowing everyone to get the total number of votes for each item without knowing what peers are voting for.

\subsection{Label Refinement Phase}

The refinement phase exploits $\mathcal{M}$ to adjust the initial predictions $\{\hat{Y}^{(p)}\}_{p=1:k_p}$.
This step is done locally where each peer will refine its own prediction $\hat{Y}^{(p)}$, without any communication.
The initial classification $y^{(p)}_0(\mathbf{x}) = A^{(p)}(\mathbf{x})$ is updated using the following equation:

\begin{equation}\label{eq:refine}
  y^{(p)}_{t+1}(\mathbf{x}) = \arg\max_{\ell \in \mathcal{L}^{(p)}} \sum_{\mathbf{x}' \in \mathcal{N}_k(\mathbf{x})} \mathcal{M}(\mathbf{x}, \mathbf{x}') \delta\left(y^{(p)}_t(\mathbf{x}'), \ell \right)
\end{equation}
In other words, $y^{(p)}(\mathbf{x})$ is updated with a weighted voter model, using $\mathcal{M}$ to adjust the neighbors' label contribution.
The update process is repeated several times until labels do not change significantly from their previous estimation.

The process can be seen as a horizontal voter model, where instead of looking at all superposed predictions $\{A^{(p)}(\mathbf{x})\}_{p=1:k_p}$,
item's nearest neighbors' label $\{A^{(p)}(\mathbf{x'})\}_{\mathbf{x}' \in \mathcal{N}(\mathbf{x})}$ are taken into account.
By re-using its prediction, the peer prevents itself from label contamination from noisy or malicious peers as only the weights can be altered but not the labels.

The weights in $\mathcal{M}$ reflect the probability of two items of being in the same class,
which indirectly encodes the ensemble's boundary location as it concerns the nearest neighbors.
Near a boundary, items on the same side have high weights, whereas items from the other side have low weights.
Therefore, only the most relevant items on the same boundary side would contribute positively.

If a boundary in $\mathcal{M}$ does not pre-exist in $M^{(p)}$, nothing would change $p$'s prediction near the boundary as an item is surrounded by items with the same label.
This case occurs when a classifier has a label definition different from the group,
which explains why we can combine classifiers with different objectives without issue.

The results of this weighting scheme  differ from the $k$-NN outcome,
where all items contribute equally or differently as in the $wk$-NN version.
When density is not homogeneous, one class may contribute more than another because items are easier to find in the neighborhood.
Therefore, items would be misclassified, whereas in our approach, relevant items are identified, allowing to rectify labels independently of the density.

\subsection{Algorithm Complexity}

The complexity of the proposed approach is decomposed as follow.
The prediction step is at least in $\mathcal{O}(n)$ per peer.
Next, the matrices $\{M^{(p)}\}_{p=1:k_p}$  are each obtained in $\mathcal{O}(k n \log n)$ to search for the $k$-NN and issue the $n \times k$ matrix.
The averaging of the matrices $\{M^{(p)}\}_{p=1:k_p}$ into $\mathcal{M}$ is made in $\mathcal{O}(k n k_p)$.
Last, the refinement step costs $\mathcal{O}(k n T)$ for $T$ iterations.
In total, each peer has a complexity of $\mathcal{O}((\log n + k_p + T) k n)$, which is larger than other ensemble methods (like boosting with $\mathcal{O}(k_p n)$), but the cost is close to be linear in $n$.
The total cost is quadratic with the number of peers as they each need to average $k_p$ matrices into $\mathcal{M}$.
This cost can be significantly reduced using \textit{Gossip} approaches by performing local averaging.

\clearpage

\section{Experimental Setup}

\subsection{Datasets}

We selected four multi-class datasets from the UCI machine learning repository \cite{Dua:2019} with a sufficient number of instances ($\sim 1,000$ per class).

A large number of instances allows us to train many classifiers on disjoint item sets while preserving a large part for testing.
Some of these datasets were already split into the \textit{training} and \textit{testing} part.
We did not keep this split and merged the two sets to obtain a larger experiment testing set.

We tested our approach over multi-class datasets because binary problems are easier to solve than multi-class problems.
Multi-class problems have a lower accuracy baseline, therefore a larger margin for improvement.
In our proposed approach, we used only positive evidences.
In a binary problem, negative evidence is equivalently informative as there are only two class possibilities.
For this particular case, $\mathcal{M}$ could be used differently to improve accuracy.

\begin{table}[h]
  \centering
  \caption{Datasets description.}\label{tab:datasets}
  \begin{tabular}{|l|r|c|c|c|}
    \hline
    Dataset  &  $n$   & $|\mathcal{L}|$ & $\text{MinClass}$  & $\text{MaxClass}$ \\
    \hline
    DryBean  & 13,611 & 7    & $3.83\%$  & $26.05 \%$\\
    PenDigit & 10,992 & 10   & $9.59 \%$  & $10.40 \%$\\
    Statlog  & 6,435  & 6    & $9.73 \%$  & $23.82 \%$\\
    USPS     & 9,298  & 10   & $7.61 \%$  & $16.70 \%$\\
    \hline
  \end{tabular}
\end{table}

Table \ref{tab:datasets} summarizes the characteristics of the different datasets.
The table gathers the number of samples $n$, the number of classes $|\mathcal{L}|$, and the proportion of the less and most represented classes $\text{MinClass}$ and $\text{MaxClass}$  respectively.
The datasets are correctly balanced, with enough samples in the lowest class, so we do not need to care about class unbalance.
We pre-processed all datasets the same way, using $z$-normalization to give an equal contribution to each feature.

\subsection{Weak Learner Setup}

In our experiments, we will test the influence of the number of classification peers in the system.
We want to train them on a disjoint subset of data to ensure they all are different.
As the datasets are limited in size, we selected the $k$-NN neighbors classifier for  simplicity and the low training data requirements.
To avoid class imbalance, each classifier's training was composed of $d$ items from each class.
$d$ was set to $3$ leading to $30$ items per classifiers for a dataset with $10$ classes like \textit{USPS} and \textit{PenDigit}.

The \textit{USPS} has the lowest ratio $\frac{n}{|\mathcal{L}|}$.
In an ensemble with $20$ classifiers, it uses $600$ items for training, representing less than $7 \%$ of the total dataset size.
For the \textit{DryBean} dataset with the largest ratio, $3 \%$ of the dataset would be used in the largest training condition.
This places the experiment under weakly supervised learning conditions, letting more space for accuracy improvement.

\clearpage

\pagebreak

\section{Results}

In this section, we present the results of the experiments comparing our ensemble method -- denoted $LR$ for \textit{Label Refinement} -- to the simple voter model -- denoted $VM$ -- using the same ensemble of classifiers.
As the $LR$ method is performed locally, not all the different peers obtain the same results.
The different predictions can be centralized and aggregated using a voter model.
This third possibility is denoted $LR+VM$ in the experiments.

\subsection{Accuracy Improvement under Stable Conditions}

\begin{figure}[h]
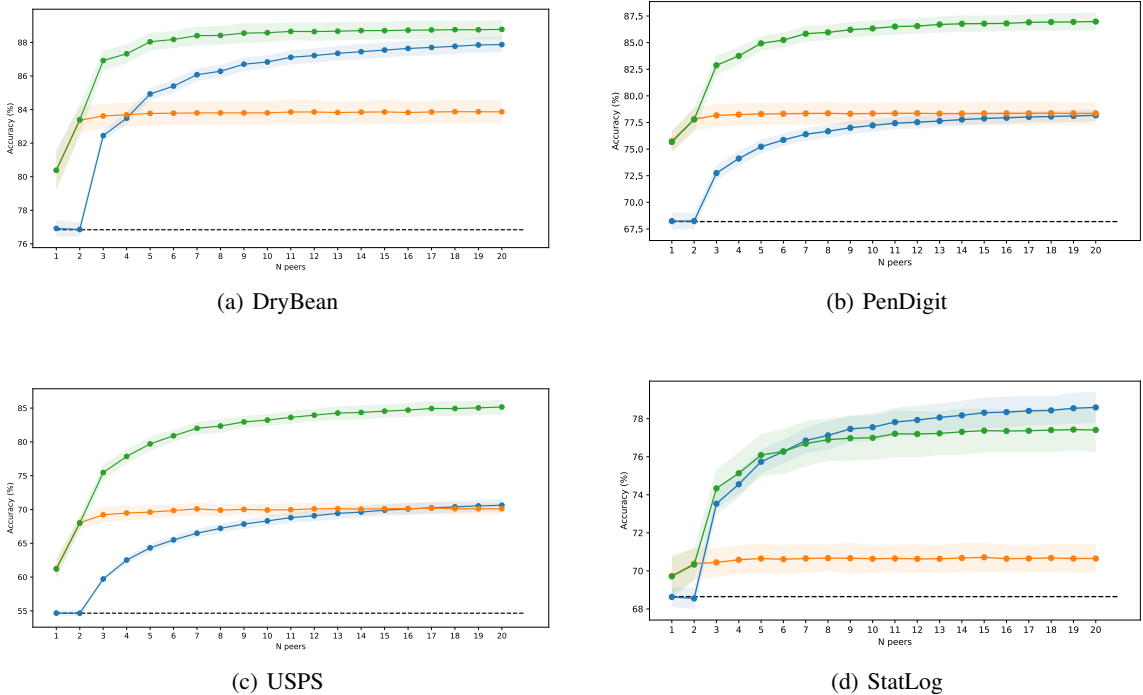

    \centering
    \def\svgwidth{0.45\textwidth}
    \subfigure[DryBean]{\input{./img/exp1/exp4_DryBean.pdf_tex}}
    \def\svgwidth{0.51\textwidth}
    \subfigure[PenDigit]{\input{./img/exp1/exp4_PenDigit.pdf_tex}}

    \def\svgwidth{0.45\textwidth}
    \subfigure[USPS]{\input{./img/exp1/exp4_USPS.pdf_tex}}
    \def\svgwidth{0.51\textwidth}
    \subfigure[StatLog]{\input{./img/exp1/exp4_StatLog.pdf_tex}}

    \caption{Accuracy evolution varying with the number of peers.
      Dashed line: average accuracy of a classifier;
      orange: $LR$;
      green line: $LR+VM$;
      blue line: $VM$.
      Shaded areas correspond to the standard deviation of the mean accuracy over 50 trials.
    }
    \label{fig:accuracy}
\end{figure}

The first experiment explores the impact of the ensemble's size on accuracy.
In this setup, the peers are assumed to be trustful and communication perfect, such as the matrices $M^{(p)}$ are exchanged without errors.
Fig. \ref{fig:accuracy} displays the results of the three different setups and the baseline corresponding to a classifier alone.

As expected, all ensemble methods are more accurate than the average classifier.
The accuracy gain of each method varies from one dataset to another, regardless of the number of classes, number of test samples or the base accuracy.

For all approaches, there is a point for $1$ and $2$ peers.
In the case of the voter model, the results correspond to the average learner accuracy as the scheme cannot help in this configuration.
For $LR$ and $LR+VM$, the first point corresponds to the mean accuracy of a learner $p$ refining its prediction with a binary matrix $M^{(p')}$ from another peer $p'$ without combining it with its own.
For $2$ peers, the internal matrix $M^{(p)}$ is combined with an external one $M^{(p')}$.
The $LR$ approach led to accuracy gain for all datasets in this configuration.

Compared to the voter model, the label refinement approach is quickly beatten over the \textit{Drybean} and \textit{StatLog} datasets.
However, it is more competitive over the \textit{PenDigit} and \textit{USPS} datasets where the voter model can only reach the same accuracy after gathering $15$ learners in the ensemble.
The accuracy gain obtained with $LR$ is quite stable after gathering just a few peers.
$5$ peers are enough to reach the maximal accuracy gain with $LR$.
In comparison, the accuracy of an ensemble using $20$ peers can still improve by adding peers under the voter model.

When applying a voter model \textit{after} the label refinement step, an additional gain of accuracy is observed.
The gain is non-negligible and allows better accuracy with $LR+VM$ over three of the four datasets with a large margin.
While the $LR$ accuracy is stable after gathering $5$ peers, the combination of $LR$ to $VM$ is beneficial as $LR+VM$ can still improve accuracy by adding more peers.
Nonetheless, it requires another communication phase to collect all the refined predictions.

\subsection{Resilience to Output Corruption}

This second experiment simulates an environment where \textit{output} labels are corrupted at random.
The corruption could happen in the peer memory or during transmission.
In the real world, it could be materialized by a truncation of the results (one part is unreadable or non-received), or by the addition of noise (some labels are flipped).
A noise level $\alpha$ corresponds to the replacement of $\alpha \%$ of the labels by a random label from $\mathcal{L}$.
A deletion  of $\alpha \%$ of the input could be considered the same way, as missing data are inputted with random labels, to the difference that the position of the incorrect label is known.

\begin{figure}
    \centering
    \def\svgwidth{\textwidth}
    \subfigure[DryBean]{\input{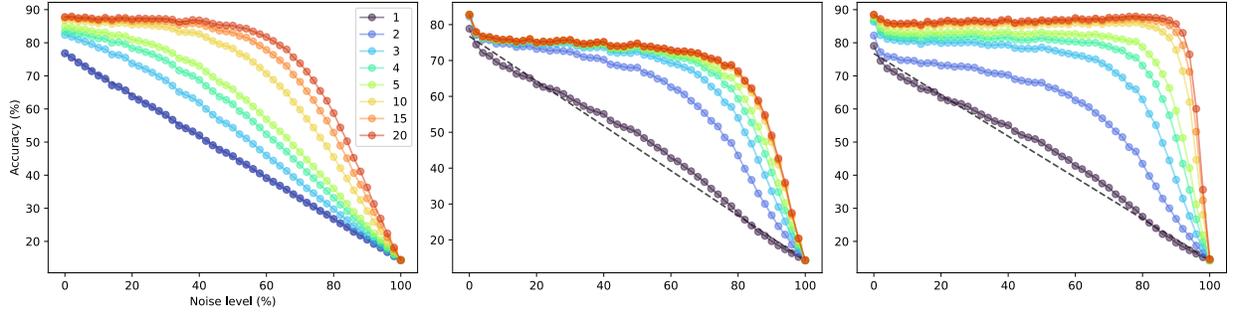}}

    \def\svgwidth{\textwidth}
    \subfigure[PenDigit]{\input{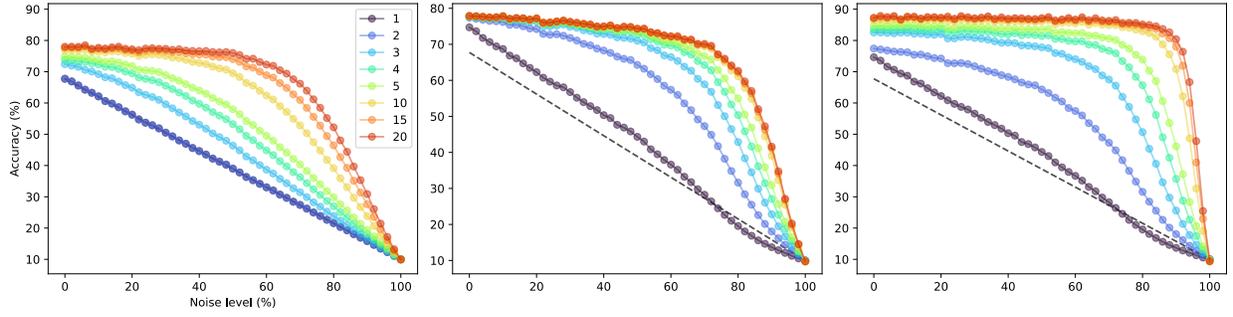}}

    \def\svgwidth{\textwidth}
    \subfigure[StatLog]{\input{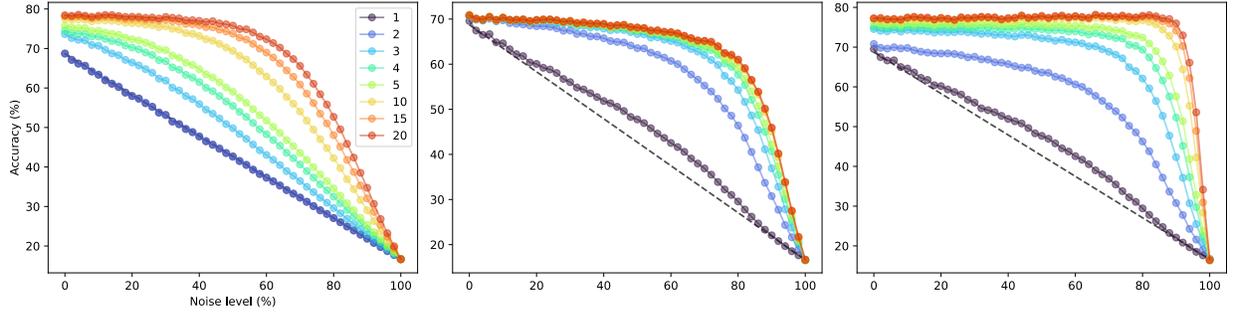}}

    \def\svgwidth{\textwidth}
    \subfigure[USPS]{\input{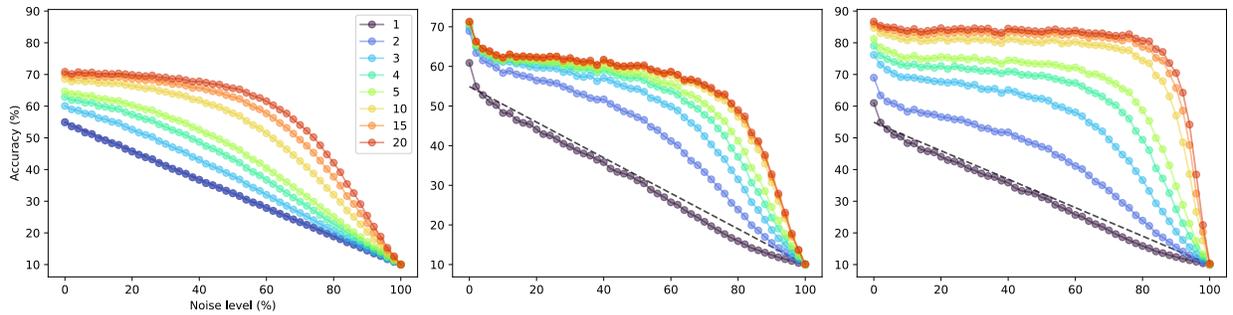}}

		\caption{Resilience to output corruption under increasing level of noise, with boundary parameter $k=10$.
    For each dataset, the ensemble models are ordered as follow: left: $VM$, middle: $LR$, right: $LR+VM$.
    Each curve corresponds to a particular ensemble size; the number of peers is indicated in the first figure's legend.
    The dashed line corresponds to the average corrupted learner accuracy.
    For $VM$, the curves corresponding to the ensemble of sizes $1$ and $2$ overlap with the dashed line.
    The noise is increased by step of $2 \%$, from $0 \%$ to $100 \%$.
    Each experiment has been repeated $25$ times.
    }
    \label{fig:noise}
\end{figure}

In this configuration, we compare the model's ability to recover the true labels $\mathcal{Y}$ using only the corrupted labels $\{\hat{Y}^{(p)}_c\}_{p=1:k_p}$.
The $\{M_c^{(p)}\}_{p=1:k_p}$ matrices are also derived from the \textit{corrupted} labels and the refinement step starts with the peer's corrupted labels $\hat{Y}^{(p)}_c$.
The results of the three different methods over the four datasets are displayed in Fig. \ref{fig:noise}.

The line corresponding to an ensemble of size $1$ corresponds -- as in the first experiment --
to the average accuracy of corrupted peers in the case of the $VM$,
and the accuracy of a peer refining its corrupted output $\hat{\mathcal{Y}}^{(p)}_c$ with a received binary matrix $M^{(p')}_c$.
When increasing the noise level, the $VM$ accuracy decreases linearly.
It is almost equal to $\text{acc}(\alpha) = (\text{acc}_0  - |\mathcal{L}|^{-1}). (1 - \alpha) + |\mathcal{L}|^{-1}$,
where $\text{acc}_0$ is the accuracy without noise, and $|\mathcal{L}|^{-1}$ the probability to select a correct label at random.
The dashed line corresponds to this equation to allow an easier comparison of the different methods.

The methods do not have the same response to an increase in noise.
$VM$ requires much more peers to obtain a resilience equivalent to $LR$.
A change from $1$ to $5$ learners offers limited improvement for $VM$, while in the case of $LR$,  an ensemble with $5$ learners is almost as stable as an ensemble of $20$ learners.
The accuracy of $LR$ using one external matrix $M_c$ for refinement does not offer resilience, as depending on the dataset and the noise level the accuracy is a little bit greater or lower than the baseline accuracy. However, moving to the use of $2$ matrices $M_c$, the resilience of $LR$ surpasses the voter model.
$LR + VM$ is even more resilient than the two other methods with very flat horizontal curves

When looking at the red curves corresponding to ensembles of size $20$, there are two observable domains: \textit{resilience} and \textit{fragility}.
In the \textit{resilience} domain, an increase of noise leads to a small decrease of accuracy, while in the \textit{fragility} domain this increase leads to an important change of accuracy.
The transition between the two domains depends on the method used and the ensemble size.
The transition occurs around $65 \%$ of noise for $VM$, nearby $80 \%$ for $LR$, and $90 \%$ for $LR + VM$.
Therefore, the label refinement approach and its variant lead to more resilience than the voter model with the same ensemble.

\subsection{Boundary Size Influence}

\begin{figure}
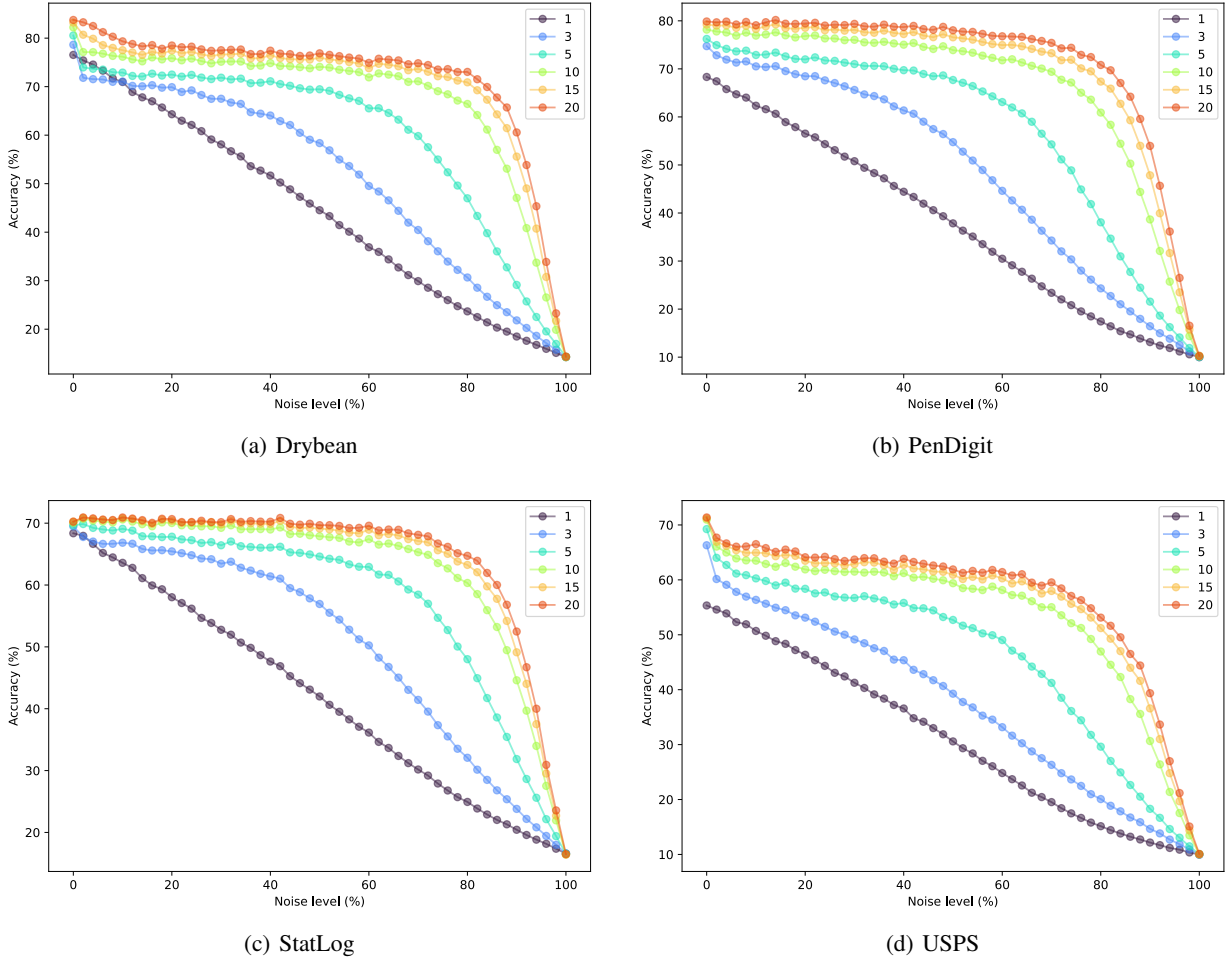

    \centering

    \def\svgwidth{0.49\textwidth}
    \subfigure[Drybean]{\input{./img/exp3/DryBean_Kbound_v_s.pdf_tex}}
    \hfill
    \def\svgwidth{0.49\textwidth}
    \subfigure[PenDigit]{\input{./img/exp3/PenDigit_Kbound_v_s.pdf_tex}}

    \vfill

    \def\svgwidth{0.49\textwidth}
    \subfigure[StatLog]{\input{./img/exp3/StatLog_Kbound_v_s.pdf_tex}}
    \hfill
    \def\svgwidth{0.49\textwidth}
    \subfigure[USPS]{\input{./img/exp3/USPS_Kbound_v_s.pdf_tex}}

    \caption{Resilience to noise  for different boundary parameters $k$,
    using $LR$ with  $10$ classifiers in the ensemble.}
    \label{fig:boundary}
\end{figure}

In the previous experiments, the boundary parameter $k$ controlling the number of nearest neighbors was set to $k=10$.
This parameter is important as it controls the number of items that can induce a label change.
Therefore, the third experiment studies the impact of $k$ over the resilience behavior for a fixed number of peers $k_p=10$.
The results are displayed in Fig. \ref{fig:boundary}.

Under the absence of noise, the increase of $k$ has a positive effect on the accuracy of the $LR$ method but is limited in amplitude.
In the presence of noise, a larger value of $k$ prevents a premature loss of accuracy.

The configuration with $k=1$ does not seem to have any impact on the accuracy.
In this setup, the only possibility for an item is to inherit from its neighbor's label unless they already have the same label.
If $\mathbf{x}$ and $\mathbf{x}'$ are reciprocal neighbors, they will exchange their labels forever.
If they are not (i.e., $\mathbf{x} \in \mathcal{N}(\mathbf{x}')$ but $\mathbf{x'} \notin \mathcal{N}(\mathbf{x})$), then $\mathbf{x}$ would take $\mathbf{x}'$'s label.
The number of possible changes in this setup is limited, which might explain the small difference from the average classifier accuracy.

The accuracy under noisy condition increase is better recovered with a larger $k$.
The switch from \textit{resilience} to \textit{fragility} also changes from $40 \sim 50\%$ for $k=3$ to $80 \sim 90$ for $k=20$.
The resilience behavior is easily obtained, with $k=10$ providing almost the same resilience to noise as $k=20$.
Compared to the curves presented in Fig. \ref{fig:noise}, an increase of the boundary $k$ in $LR$ leads to greater resilience in accuracy than the addition of more peers in the $VM$.
A sufficient $k$ needs to be combined to a sufficient $k_p$ to benefit from the ensemble size and items' neighborhood stability.

The greater resilience to noise with a larger $k$ can be simply explained.
In a dense area where all items belong to the same class (before corruption), all nearest items would receive the same weight.
Therefore the weighted vote will be equivalent to a normal vote.
As $k$ increases, more intact labels would be included in the vote, making the decision  more stable.
Consequently, items in this area will easily recover their initial labels.
The items near the boundary benefit from the same effect because they are surrounded by boundary items that are possibly misclassified.
Therefore, $k$ needs to be larger for these area to recover.

When $k$ is small, the density is almost invariant; therefore, all classes might be represented in equal proportion.
However, the density can vary from one location to another within the radius when extending the radius.
If located in a very dense area, a class will be more numerically represented.
The vote in \eqref{eq:refine} is biased by the weights, but also by the number of items that contribute.
As an example, if there is one item $\mathbf{x}'$ in the neighborhood of $\mathbf{x}$ belonging to the correct class $\ell$ with weight $\mathcal{M}(\mathbf{x}, \mathbf{x}') = 1$,
but $10$ other items of the same incorrect class $\ell_w$  with weight $\mathcal{M}(\mathbf{x}, \mathbf{x}') = 0.11$, then $\ell_w$ would win the vote.
Therefore, the parameter boundary $k$ must be chosen appropriately to avoid this type of issue. In addition, the process becomes computationally expensive.

\clearpage

\section{Discussion}

\subsection{Instance vs Batch Classification}

One major difference between our ensemble method and classical ensemble methods is the input size.
Traditionally, the ensemble process one sample at a time, allowing us to predict labels of any sample size without constraint.
Our approach is inspired by \textit{Evidence Accumulation Clustering}, where samples are processed by batch.
Therefore, it limits the usability of  use-cases that do not require real-time answers.
Nonetheless, for applications with no real-time constraint, data can be stored into a buffer and classified when the collected data is sufficient.

In the different experiments, we classified all the items left in the test set together.
When reducing the test size, the system can become unstable if too few samples are available.
In a batch, items can be classified into \textit{core} items that are easily classified by all peers, with most of their nearest neighbors belonging to the same class,
and \textit{boundary} items where errors are more frequent.
Misclassified core items are easily rectified thanks to their strong neighborhood.
However, boundary items rely on core items to get good reference labels, as other boundary items may not be reliable enough.
If the batch size is too small, core areas would be restricted to very few points, and would not be large enough to offer stable references to other items.

We tested over the \textit{DryBean} dataset to reduce the batch size.
Before reaching the system instability for $k=10$ and its $7$ classes, the critical size limit was $150$ items.
Under this size, the mean accuracy decreased while the variance increased.

One way to work on smaller batches is to adapt the boundary parameter $k$.
Decreasing its value may help to keep core areas stable over small batch sizes, while a too large $k$ might destroy core areas due to misclassified boundary items.
Another strategy to work on small batches is to classify the batch $X$ together with $X'$ obtain from a database.

The full process is performed on $X_0 \cup X_1$, which would not suffer from possible instabilities.

\subsection{Combining Clustering with Classification}

The use of co-association matrices $\{M^{(p)}\}_{p=1:k_p}$ rather than raw predictions $\{\hat{Y}^{(p)}\}_{p=1:k_p}$
gives the possibility to adapt our approach to clustering or mix clustering with classification results.
The combination of the two types of algorithms may help to improve classification accuracy.
However, the clustering's benefits might be more limited.

One of the main clustering problems is the selection of the number of clusters.
Unfortunately, ensemble clustering algorithms often cluster $\mathcal{M}$ using a hierarchical algorithm that circumvents the cluster number problem.
The label refinement scheme will help adjust the boundary but will neither help find the best number of clusters.

A direction to explore is how to fuse (or split) clusters using ensemble information.
A possibility would be to search for clusters boundaries looking at the local CA matrix $M^{(p)}$ that does not exist in the ensemble matrix $\mathcal{M}$.
By detecting weak cluster boundaries, the clusters separated by these boundary could be merged.
The same reasoning could be applied for splitting clusters by searching for a strong boundary in $\mathcal{M}$ that does not exist in $M^{(p)}$.
This direction can be explored and compared to a more direct clustering of $\mathcal{M}$.

\subsection{Ensemble of $n$-ary Classifiers}

Mutli-class classifiers can be built by combining binary classifiers, as in the ECOC \cite{Dietterich1995SolvingML}.
Each classifier is trained over $2$ classes obtained by splitting the $d_0$ initial classes at random.
This class grouping exploits \textit{class synergies} as two similar classes are easier to distinguish from the other classes when grouped than each alone.
The original multi-class would be identified by looking at the ensemble classification binary code obtained.

More generally, $d_0$ classes can be fused into $d_1$ classes.
However, the transformation of the multi-class prediction to fewer classes can  obfuscate the results for privacy reasons.
Another motivation is the reduction of the transmitted bit when exchanging the raw predictions.

If $d_0 \gg d_1$, the amount of preserved boundaries is $1 - \frac{1}{d_1}$ (the boundary between two classes merged into the same group is no more identifiable, but still observable for two classes in a distinct group).
It is at worse $50\%$ for a binary grouping, impacting $\mathcal{M}$ in these proportion.

Looking at particular class boundaries, a proportion $\alpha$ of the classifiers still see this boundary and would vote something different for each pair of items.
On the other hand, a proportion $(1-\alpha)$ do not see it and would vote $1$ for any pairs.
Therefore, we could rewrite $\mathcal{M}$ as $\mathcal{M}(\mathbf{x},\mathbf{x}') = \alpha m(\mathbf{x}, \mathbf{x}') + (1-\alpha)$ where $m(\mathbf{x}, \mathbf{x}')$ is the average co-association score of the classifiers seeing the boundary.
When performing the vote using Eq. \eqref{eq:refine}, the part $(1-\alpha)$ contributes equally for all neighbors leading to a very limited effect on the vote outcome.

\section{Conclusion}

In this article, we proposed a noise-resilient ensemble classification method exploiting the co-classification of nearest items.
The system works in two steps: an \textit{exchange phase} where peers communicate their prediction, and a local \textit{label refinement phase} where labels are adjusted using a weighted voter model.

This ensemble approach was tested on four different datasets, where it led to accuracy improvement.
Moreover, under the noisy condition, the proposed ensemble method was highly resilient to noise.
It allowed it to preserve accuracy with much fewer peers than an ensemble combining its prediction using a voter model.
The approach is flexible, as clustering peers can participate in helping classification peers.
Additionally, classifiers with different objectives can be combined if some of their class shares the same boundaries.

In future works, we will analyze the data factors impacting the accuracy to quantify the possible gain,
and investigate how to estimate the boundary parameter $k$ to better adapt to the dataset to classify.

\clearpage

\pagebreak

\bibliographystyle{unsrtnat}

\bibliography{references}  

\end{document}